\documentclass[runningheads]{llncs}
\usepackage{graphicx}
\usepackage{amsmath,amsfonts}
\usepackage{cite}
\usepackage{url}
\usepackage[ruled,vlined]{algorithm2e} 
\usepackage{booktabs}
\begin{document}
\title{TCR: Short Video Title Generation and Cover Selection with Attention Refinement}
\titlerunning{Short Video Title Generation and Cover Selection}
%
\author{Yakun Yu\inst{1}\thanks{Equal contribution.} \and
Jiuding Yang\inst{1}$^\star$ \and
Weidong Guo\inst{2}$^\star$ \and
Hui Liu\inst{2} \and
Yu Xu\inst{2} \and
Di Niu\inst{1}}
\authorrunning{Y. Yu et al.}
%
\institute{University of Alberta, Edmonton, AB, Canada\\
\email{\{yakun2,jiuding,dniu\}@ualberta.ca}
\and
Tencent, Shenzhen, China
}
\maketitle              
\begin{abstract}
With the widespread popularity of user-generated short videos, it becomes increasingly challenging for content creators to promote their content to potential viewers. Automatically generating appealing titles and covers for short videos can help grab viewers' attention. Existing studies on video captioning mostly focus on generating factual descriptions of actions, which do not conform to video titles intended for catching viewer attention. Furthermore, research for cover selection based on multimodal information is sparse. These problems motivate the need for tailored methods to specifically support the joint task of short video title generation and cover selection (TG-CS) as well as the demand for creating corresponding datasets to support the studies. In this paper, we first collect and present a real-world dataset named Short Video Title Generation (SVTG) that contains videos with appealing titles and covers. We then propose a Title generation and Cover selection with attention Refinement (TCR) method for TG-CS. The refinement procedure progressively selects high-quality samples and highly relevant frames and text tokens within each sample to refine model training. Extensive experiments show that our TCR method is superior to various existing video captioning methods in generating titles and is able to select better covers for noisy real-world short videos.

\end{abstract}
\section{Introduction}
Video titles and covers are two critical elements for capturing users' attention when using an app (e.g., TikTok).
However, many short videos do not have well-edited titles beyond plain descriptions or a cover image other than the first frame. 
The sheer volume of daily video uploads further makes it hard for the platform to edit video covers in a timely manner.
Therefore, there is an apparent demand for using artificial intelligence to improve the quality of video titling and cover image selection, especially with the objective of increasing user engagement.

Various video description generation techniques have been studied and developed recently, including video captioning  \cite{iashin2020multi, tan2021learning}, video-based summarization  \cite{li2020vmsmo, khullar2020mast}, etc. 
They are mainly designed for objectively describing actions and interactions of objects in a video.
For example, for a video of a woman doing yoga, a generated caption is usually ``a woman is exercising while her cat keeps disturbing her'', which is correct but unvarnished. However, a video uploader would rather title the scene as ``My cat sees me as his toy'' in a hilarious way in order to attract viewers' attention.
Therefore, there is still a gap in the literature to generate appealing and human-like video titles.
However, a lack of open-sourced short video datasets from real-world social media platforms hinders the development of techniques for human-like video title generation. Furthermore, little work has been done on video cover selection \cite{ren2020best}, which is another challenge, again due to the lack of video cover selection dataset. 

To address the issues mentioned above, we collect a Short Video Title Generation (SVTG) dataset that consists of 8,652 short video samples with a human-edited appealing title and a visual cover selected by the original content creator. To the best of our knowledge, SVTG is the first publicly available dataset that is designed for joint video title generation and cover selection (TG-CS) on real-world short videos, especially with a purpose of boosting click-throughs in mind rather than plainly describing the scene.

Furthermore, we propose a novel video Title generation and Cover selection with Refinement (TCR) approach that integrates a title-cover generator with cross-attention-based refinement learning for TG-CS task based on the multimodal information in short videos. 
Specifically, 
we use a multimodal title-cover generator to capture the temporal dependency between modalities. In order to handle noisy modality data extracted from real-life short videos, we further propose a refinement learning strategy to refine model training. 

Our main contributions are summarized as follows:
(1) We construct a new short video dataset, SVTG, which can facilitate research on short video title generation beyond generating the unvarnished factual description of objects, and research on video cover selection based on multimodal information. Our objective is to generate eye-catching video titles and covers, with the goal of engaging online user attention;
(2) We propose the TCR method that can simultaneously generate appealing titles and covers by fully capturing the dependency between modalities and refinement training;
(3) We evaluate TCR on the SVTG dataset as well as on the public How2 dataset.
TCR has demonstrated its effectiveness on title generation across the two datasets and on cover selection under SVTG. 
We will open-source the dataset and code to facilitate future research on generating titles and selecting covers for real-world social media videos.
\section{Related Work}
\subsection{Text Generation}

Text generation mainly includes text summarization and video captioning. The former aims to generate a short summary for a long document while preserving the main idea of the document. The latter focuses on describing the visual content of a video using condensed text.

Conventional text summarization utilizes a single text modality to generate summaries \cite{wang2020heterogeneous, nallapati2017summarunner}.
Recent works \cite{li2020vmsmo, khullar2020mast} show promising results on summarization with the help of other modalities, e.g., the visual modality.  
For instance, 
Liu \textit{et al.} \cite{liu2020multistage} propose both RNN-based and Transformer-based encoder-decoder models with the input of videos and its ground-truth/ASR transcripts.
Encoder-decoder networks are widely used in video captioning where the encoder converts the input into high-level contextual representations before the decoder takes the representations and generates words as the caption. For example,  
Tan \textit{et al.} \cite{tan2021learning} propose a RNN-based encoder-decoder model with three visual reasoning modules to generate video captions.

The above methods perform well on generating factual descriptions of a video scene given the video and the corresponding well-edited transcripts, yet fail when they intend to generate titles beyond factual descriptions based on noisy multimodal information. In contrast, our method can handle noisy modality data and capture the subtle relations between modalities, thus being able to generate both factual descriptions and appealing titles.

\subsection{Cover Selection} 

Cover selection aims to find a representative frame that conveys the main idea of a video. Song \textit{et al.} \cite{song2016click} propose a thumbnail selection system using clustering methods. Ren \textit{et al.} \cite{ren2020best} select the best frame based on the prediction scores. They utilize only the visual modality, which is insufficient for real-world short videos where the keyframe should be determined by multiple modalities.  
Though Li \textit{et al.} \cite{li2020vmsmo} propose VMSMO, to train a cover selection model based on multimodal information, i.e., news video and news article pairs, 
their model's performance is highly dependent on the quality of news articles that are well-edited by humans. However, the SVTG dataset does not provide such manually-created articles. Therefore, it is inappropriate to apply their method to our task for comparison. Moreover, we extract covers using a unified model instead of an independent cover selection model, which is more straightforward.

\section{The SVTG Dataset}

\begin{table}[t]
\caption{The statistics of the SVTG dataset.\label{tab:stat}}
\centering
\begin{tabular}{lccccc}
\toprule
 & \#Videos & \#Categories  & \#Authors & \#Sentences  &Vocab  \\ \midrule
Train   & 8,052      & 37        & 3,478      & 155,826     & 100,361   \\
Valid   & 200       & 25        & 172       & 3,802       & 6,770   \\
Test    & 400       & 27        & 288       & 7,352       & 10,494   \\
All     & 8,652      & 37        & 3,618      & 166,980     & 105,487  \\ 
\bottomrule
\end{tabular}

\end{table}

To the best of our knowledge, SVTG is the first publicly available Chinese dataset that designed for the joint TG-CS for short videos.
We regard a title as appealing if it expresses the video maker's desire to attract viewers with some sentiment words in addition to describing the factual information (e.g., the interactions between objects). See Supplementary A for SVTG data collection.
Table \ref{tab:stat} shows the statistics of our dataset.

\begin{figure*}[!t]
	\centering
		    \includegraphics[width=0.95\textwidth]{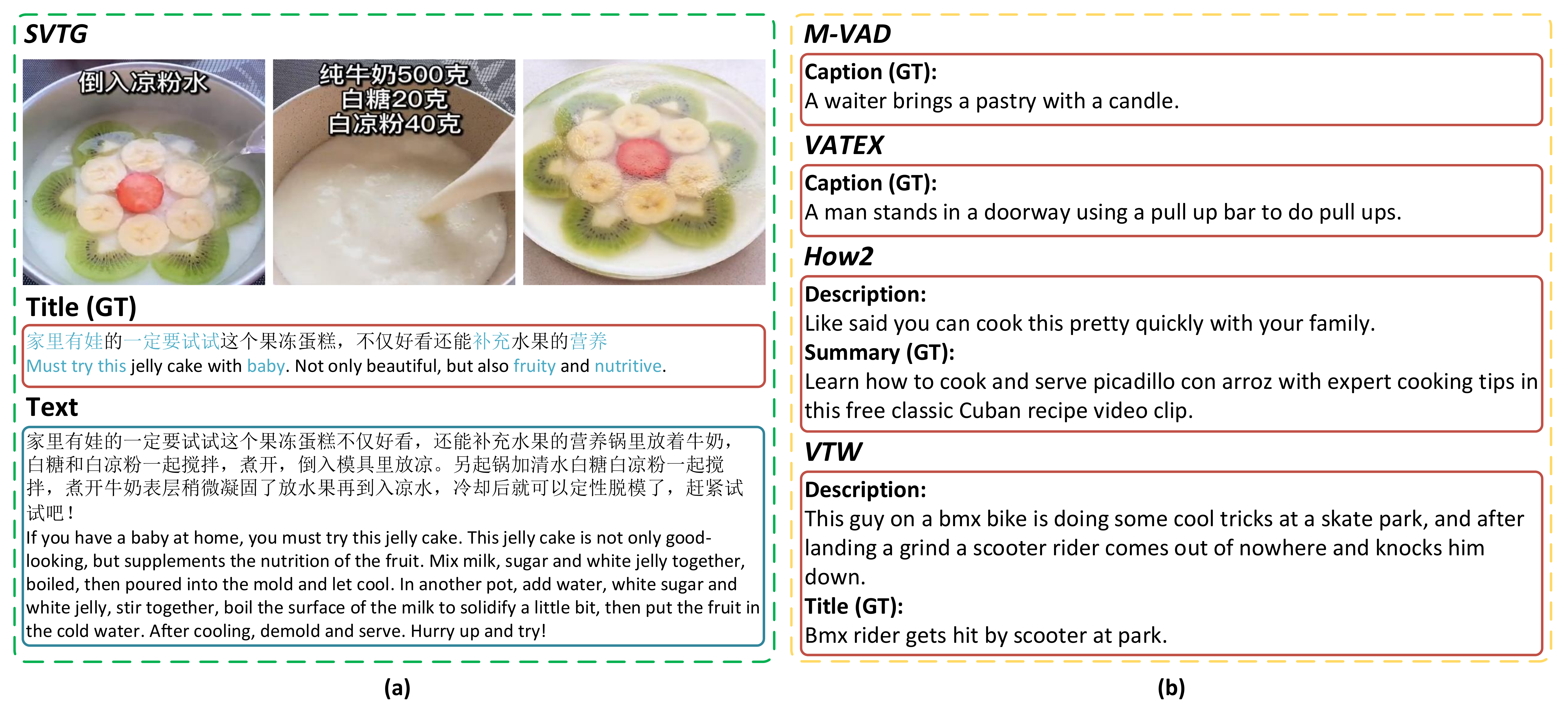}
		\caption{Comparison between (a) SVTG data example and (b) Other benchmark dataset examples. SVTG example has the continuous frames on the top, title on the middle, and transcripts extracted by ASR/OCR on the bottom. Other examples show their ground truths (GT) and user-generated descriptions.}
	\label{fig:sample}
\end{figure*}
To show the uniqueness of SVTG, we compare it with four other benchmark datasets, as shown in Fig. \ref{fig:sample}. 
Compared with these datasets, SVTG has three unique characteristics.
First, SVTG videos have more appealing titles, which can promote the study of deep learning models to generate short video titles beyond plain factual descriptions.
As we observe in Fig. \ref{fig:sample}(a), SVTG videos usually contain strong sentiment words (e.g., ``must'' and ``nutritive'') to draw viewers' attention except simply describing the video scene. 
In contrast, from Fig. \ref{fig:sample}(b), we can see that the captions, summaries, and titles in other benchmark datasets usually emphasize on the actions of objects (e.g., ``stand'') and the interaction between objects (e.g., ``a man using a bar''). These descriptions tend to be unvarnished and objective without any sentiments. 
If we annotated the video from Fig. \ref{fig:sample}(a) in the way as the other datasets do, the title would sound like ``pour milk and add fruit to make cake'' and would be less attractive to online viewers. 

Second, SVTG is more challenging for video title generation, and closer to real-life videos. 
As demonstrated in Fig. \ref{fig:sample}(b), the existing benchmark datasets contain human-annotated video descriptions, subtitles or articles as the text modality.
However, it is expensive or even infeasible to have such human-annotated video descriptions in real-world applications.
In real-world settings, when a content creator uploads a short video to the platform, the platform needs to select a cover and generate a title based on the video itself.
All information in SVTG comes from the video itself without any other human annotations. Thus SVTG is closer to real-world applications.
However, ASR- and OCR-detected text can be highly noisy in contrast to human-annotated summaries or descriptions, making SVTG more challenging than other datasets. 

Third, to our best knowledge, SVTG is the first dataset that includes the video covers for facilitating the joint TG-CS on short videos. We argue that the combination of two tasks, title generation and cover selection, is crucial for attracting viewers' attention and thus should not be separated for consideration. Furthermore, having a unified dataset encourages the model to learn the shared connection of this joint task.
See Supplementary B for more details about these benchmark datasets.

\section{Method}

\begin{figure*}[!t]
	\centering
		    \includegraphics[width=\textwidth]{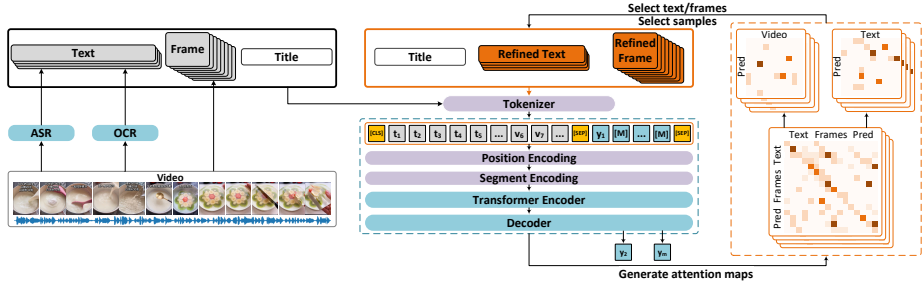}
		\caption{The overall architecture of the proposed TCR method including data preprocessing in the black box, a title-cover generator in the blue dotted box, and a multimodal cross-attention-based refinement module in the orange dotted box. ``Title'' means the ground truth title. ``Pred'' in the attention map represents the predicted title. And ``[M]'' in the tokenized sequence stands for a masked token.}
	\label{fig:framework}
	\vspace{-3mm}
\end{figure*}

We propose a self-attentive method named TCR for the joint TG-CS under the SVTG dataset. Fig. \ref{fig:framework} shows the structure of the TCR method.
It consists of a multimodal title-cover generator and a refinement module. 
The generator generates titles and extracts covers by fully exploiting the dependency between modalities.
The refinement module helps refine model training by selecting high-quality samples and highly relevant tokens/frames within each sample.
We will explain the TCR method in detail below.

\subsection{Title Generation and Cover Selection}
Since the frames and the text inputs are both temporally ordered, we can directly combine them to explicitly learn the dependency between all pairs of text tokens and frames. We first preprocess the training data into the following formats:

\begin{itemize}
    \item {Text Representation:} given the transcripts extracted by ASR and OCR from a video, the combined ASR and OCR text is tokenized into a sequence of WordPieces \cite{wu2016google}, i.e, $T = (t_1, t_2, ..., t_n)$, where $n$ is the number of tokens in the sequence. 
    \item {Title Representation:} the title is represented as a sequence of tokens $Y = (y_1, y_2, ..., y_m)$, where $m$ denotes the number of tokens in $Y$.
    \item {Video Representation:} given a sequence of video frames of length $L$, we feed it into a pre-trained ResNet-101 2D convolutional neural network \cite{he2016deep} to obtain visual features $V = \{v_l\}_{l = 1}^L \in \mathbb{R}^{L \times d_v}$, where $d_v$ is the feature dimension of the pretrained ResNet model.
\end{itemize}

The title-cover generator is composed of three embedding layers, 12-layer transformer encoders, and a language modeling (LM) head on top as the decoder.
On the one hand, we regard the frames as $L$ virtual tokens $\{v_i\}_{i = 1}^L$ and concatenate them with text tokens $T$ to form a new sequence of length $L + n$. We also add a special token [CLS] at the start of the sequence and another special token [SEP] at the sequence end. These tokens, including Y, are embedded by three embedding layers that are trainable: token-level embedding, position-level embedding, and segment-level embedding.  

On the other hand, we feed the frame features $V$ into a fully connected (FC) layer to project it into the same lower-dimensional space as the text token embeddings and regard the projected frame features as the frame token embeddings.
To establish a connection between the text tokens and frames so that they could attend to each other, we combine their respective embeddings by simply replacing the virtual token embeddings with the frame embeddings. Therefore, the final representations for text tokens and video frames can be obtained by summing up their corresponding three embeddings. We then feed these representations as the input to the transformer encoders.

Assume the input is $X_{VT} \in \mathbb{R}^{L' \times d_h}$ where $d_h$ is the hidden size of our model, the output of the \textit{i}-th transformer encoder is denoted as $O_i \in \mathbb{R}^{L' \times d_h}$. Each encoder aggregates the output of the previous encoder using multiple self-attention heads, thus $O_i$ is computed by:
\begin{equation*}
    O_i = 
    \begin{cases}
    X_{VT},                         & i = 0;\\
        \text{Encoder}_i (O_{i-1}),   & i > 0.
    \end{cases}  
    \label{equ:trans_output}
\end{equation*}
In this way, we can obtain visual-language fusion representations at different encoder layers. Let $A_i$ denotes the attention score of each head in the \textit{i}-th encoder. 
\begin{equation}
    \label{equ:atten}
\begin{aligned}
        A_i = \text{Dropout} (\text{Softmax} &(\frac{Q_i K_i^T}{\sqrt{d_h/12}} + \text{Attention Mask})),\\
    Q_i, \ K_i= & \ O_{i-1}W_{Q_i}, \ O_{i-1}W_{K_i}.
\end{aligned}
\end{equation}
$Q$ and $K$ are the queries and keys linearly transformed from the input of each encoder. $\text{Attention Mask} \in \mathbb{R}^{L' \times L'}$ ($\text{elements} \in \{0, -\infty\}$) is used to make sure the target token to be generated only attending to the leftward information including itself, the generated tokens, frames and text tokens \cite{dong2019unified}. 
The LM head consists of two dense layers and a layer normalization (LN) layer in between to calculate the vocabulary distribution for decoding. 
Once the generator is well-trained, we select a frame with the highest attention score obtained in Equation \ref{equ:atten} as the video cover.

\subsection{Attention-based Model Refinement}
Considering redundant noise may exist in the ASR/OCR transcripts and continuous frames, noise filtering is intuitive to help improve the quality of the generated titles. To this end, we propose an attention-based refinement module, which 
will automatically select not only the key parts of the multimodal input within each sample but also select higher-quality samples. Moreover, the module can interact with the above generator to find the best combination of different input tokens for progressively refining the generator.

\textbf{Token-level refinement.} 
Given a title-cover generator, we could first generate titles for the training data. We then acquire the cross-attention scores between the generated titles and the inputs, i.e., the ASR/OCR tokens and the frame tokens, through Equation \ref{equ:atten}. 
We assume the tokens that the generated titles attend to a lot are critical for building a generator with good performance. Therefore, we extract them as refined sentences/frames. 
For sentence refinement, we first locate the sentence that has the highest attention score with each generated token. Then, the top $u$ sentences which the generated title attends to frequently are naturally selected as the key sentences.

Video titles always contain some words (e.g., ``alley-oop'' in a basketball game video) that frequently appear in the ASR/OCR text but are rarely related to video frames. Therefore, we need to select the related frames at a higher granularity than the way we select related sentences.
To achieve this, we first assign a $weight \in [0, L]$ to each frame according to its attention scores with each generated title token, then calculate the total weights for each frame. Higher weights indicate greater contributions to title generation. We finally pick out the top $v$ frames with the highest weights as the refined frames or keyframes for the video.

\textbf{Sample-level refinement.} In addition to refining the tokens within each sample, we denoise the data by selecting higher-quality data at the sample level. A training data will be deemed high-quality if the generated title by the generator $M$ is similar to its ground truth title. The similarity is computed by the Rouge score \cite{lin2004rouge}. 

Apparently, with refinement at both token-level and sample-level, we could refine the title-cover generator multiple times until we get the best test results. See the detailed refinement algorithm in Supplementary C.

\section{Experimental Results}
\label{sec:exp}

\subsection{Baseline Models}
We compare our proposed method with the following baseline models:

\textbf{Lead-3} directly selects the first three sentences from the text as the title \cite{nallapati2017summarunner}. 

\textbf{HSG} constructs a heterogeneous graph with sentence nodes and word nodes, then selects sentences as the summary for a document by node classification \cite{wang2020heterogeneous}. 

\textbf{MAST} is a model consists of encoder layers, a hierarchical attention layer, and a decoder that generates a textual summary on multimodal inputs \cite{khullar2020mast}.
 
\textbf{NMT} is an RNN-based model \cite{caglayan2017nmtpy} designed for sequence-to-sequence tasks and modified for video description tasks. 

\textbf{MFN} generates a summary based on video and its ASR or ground-truth transcript through a multistage fusion model with a forget gate module to remove the redundant information \cite{liu2020multistage}. 

\textbf{Bert2Bert} contains a text BERT encoder, a video transformer encoder and a BERT decoder, which is a popular framework used in recent video/image + language works \cite{hu2021unit}.

\subsection{Implementation Details}
\label{subsec:Implem}
We evaluate the above baseline models and our TCR model on the SVTG dataset following the split criteria in Table \ref{tab:stat}. We implement our experiments using PyTorch \cite{paszke2019pytorch} on an NVIDIA V100 GPU. To preprocess the videos, we sample 25 frames for each video and extract visual features from these sampled frames.

Our multimodal generator is adapted from a text pretrained model architecture \cite{dong2019unified}. 
The vocabulary size is 21,128. We use AdamW optimizer with a learning rate of $1e-5$. Linear warmup schedule is set for the first $10\%$ of the total training steps with linear decay. Dropout is adopted for regularization. The batch size is set to 16, and the maximum input sequence length is set to 512. During training, we only mask $20\%$ of the title tokens, and the maximum number of masked tokens is set to 20. We 
train up to 20 epochs {with the cross-entropy loss computed by the predicted title tokens and the ground truth tokens}. The validation set is used to select the best model with the highest mean score of the evaluation metric. For testing, we use beam search with beam size 5 to report the final test results. For refinement, we select the top 3 sentences/frames and perform one iteration.

In addition, we evaluate the TCR on the publicly available How2 dataset \cite{sanabria2018how2} to further validate its generation ability. The How2 dataset consists of 79,114 videos accompanied by corresponding user-generated descriptions and summaries. It has a train set of 73,993 samples, a validation set of 2,965 samples, and a test set of 2,156 samples. Since the TCR method is designed for handling real-world short videos with noisy multimodal information, we use the ASR transcripts instead of the provided descriptions following the work of Liu \textit{et al.} \cite{liu2020multistage} for a fair comparison.

\subsection{Results}
Here we first report the performance of the baseline models and our proposed TCR for title generation. Then we show the evaluation results of cover selection. Later, we present the ablation tests from different aspects. 

\begin{table}[t]
\caption{Rouge score comparison with the baselines for SVTG.\label{tab_result}}
\centering
\begin{tabular}{lccc}
\toprule
Model                       & R-1                 & R-2                & R-L    \\ \midrule
Lead-3                &29.44      &18.40      &37.35      \\
HSG               &41.06      &27.45      &40.59      \\ \midrule
NMT              &41.20      &28.74      &41.91      \\
MFN-rnn        &16.64      &1.47      &14.36      \\
MAST           &39.73      &27.42      &41.00          \\ 
MFN-transformer                 &19.24      &3.99      &18.40       \\
Bert2Bert      &24.84         &5.75       &22.58             \\ \midrule
TCR w/o Text         &8.53           &0.31           &9.30              \\
TCR w/o Visual        &46.35      &32.67      &46.07         \\ 
TCR                 &\textbf{47.62}      &\textbf{33.98}      &\textbf{47.08}                      \\ \bottomrule
\end{tabular}

\end{table}

\textbf{Title generation performance.} 
We evaluate the quality of the generated titles by the standard Rouge F1 matric \cite{lin2004rouge} following previous works \cite{wang2020heterogeneous, khullar2020mast, liu2020multistage, nallapati2017summarunner, li2020vmsmo}. R-1, R-2, and R-L refer to the unigram, bigram, and longest common sub-sequence overlap with the ground truth title, respectively.

We first compare the results of our method and the baseline models on SVTG, as listed in Table \ref{tab_result}. 
We can observe from the table that TCR outperforms all the baseline models in terms of all Rouge scores. Specifically, TCR outperforms the best baseline model by 6.42 R-1 points, 5.24 R-2 points, and 5.17 R-L points, indicating its superior ability to learn video titles when given noisy ASR/OCR text and frames. 
Though these baseline models perform well in generating summaries and captions given user-annotated text descriptions, they are incapable of generating titles in social media language and handling noisy text information. In contrast, TCR can generate more accurate titles by fully exploiting the dependency between all text/frame tokens and denoised training.

\begin{table}[ht]

\caption{Rouge score comparison for How2. \label{how2_re}}
\centering
\begin{tabular}{lccc}
\toprule
Model                       & R-1                 & R-2                & R-L    \\ \midrule
S2S                &48.1      &28.2      &43.4      \\
FT               &51.1      &31.0      &45.8     \\ \midrule
HA-rnn             &53.9      &34.2     &48.7     \\
HA-transformer        &55.1     &36.0     &50.1      \\
MFN-rnn                 &60.0      &43.6      &56.1       \\
MFN-transformer      &59.3        &42.1       &55.0           \\ \midrule
TCR                 &\textbf{60.9}      &\textbf{44.2}      &\textbf{60.6}                      \\ \bottomrule
\end{tabular}

\end{table}

To further examine the generation ability of TCR, we apply it to a public dataset, How2, that is designated to generate a factual summary for videos. Table \ref{how2_re} presents the results on How2. We use the dataset and baseline models in the work \cite{liu2020multistage} for a fair comparison. 
See Supplementary D for model details about these baselines. 
We can observe from Table \ref{how2_re} that TCR outperforms the baseline models in all Rouge scores. Notably, it outperforms the best baseline MFN-rnn model by 4.5 R-L points. This finding again indicates that TCR is excellent at generation both within factual descriptions (e.g., summaries in How2) and beyond factual descriptions (e.g., titles in SVTG). 
Moreover, by comparing the performance of MFN in Table \ref{tab_result} and Table \ref{how2_re}, we can conclude that MFN is unable to generalize to SVTG that is distinct from traditional video summary/caption datasets.   

\textbf{Cover selection performance.} 
We compare the selected cover images with the default cover that comes with the videos by human judgments from the following aspects
: \textit{Informativeness} ({which picture corresponds to the topic conveyed by the video title}), and \textit{Superior} (which picture is better to be a video cover that attracts you). See Supplementary E for more details about the human evaluation.

\begin{table}[t]
\caption{Evaluation scores on the selected frames and the default covers that come with the videos.\label{cover_result}}
\centering
\begin{tabular}{lcc}
\toprule
Pictures                       & \textit{Superior}                 & \textit{Informativeness}                    \\ \midrule
Our selected frames               &31.5\%      &15.5\%            \\
Default covers               &25.5\%      &15.5\%            \\
Cannot tell              &43.0\%      &69.0\%            \\ \bottomrule

\end{tabular}

\end{table}

The results are reported in Table \ref{cover_result}. We can observe from the table that participants cannot tell which one is better in most cases because the two pictures are almost identical. Moreover, our selected covers outperform the default covers by 23.5\% in terms of \textit{Superior} while they are equivalent in \textit{Informativeness}, which demonstrates the effectiveness of our method to automatically select the cover image from a series of frames.

\textbf{Ablation studies.}
All ablation studies are performed on the SVTG dataset. The last three rows in Table \ref{tab_result} show the ablation results on the effect of different modalities.
From this table, we can see that TCR outperforms TCR w/o Visual and TCR w/o Text, which demonstrates that multimodal information can benefit title generation more than a single modality.
It can also be easily observed that the text modality is dominant while the visual modality has less contribution in title generation because ASR- and OCR-detected text usually contains words that describe the highlights of stories in short video cultures. Supplementary F further provides the ablation tests on the number of selected sentences/frames and refinement iterations.

\subsection{Qualitative Analysis}

\begin{figure*}[tp]
	\centering
		    \includegraphics[width=\textwidth]{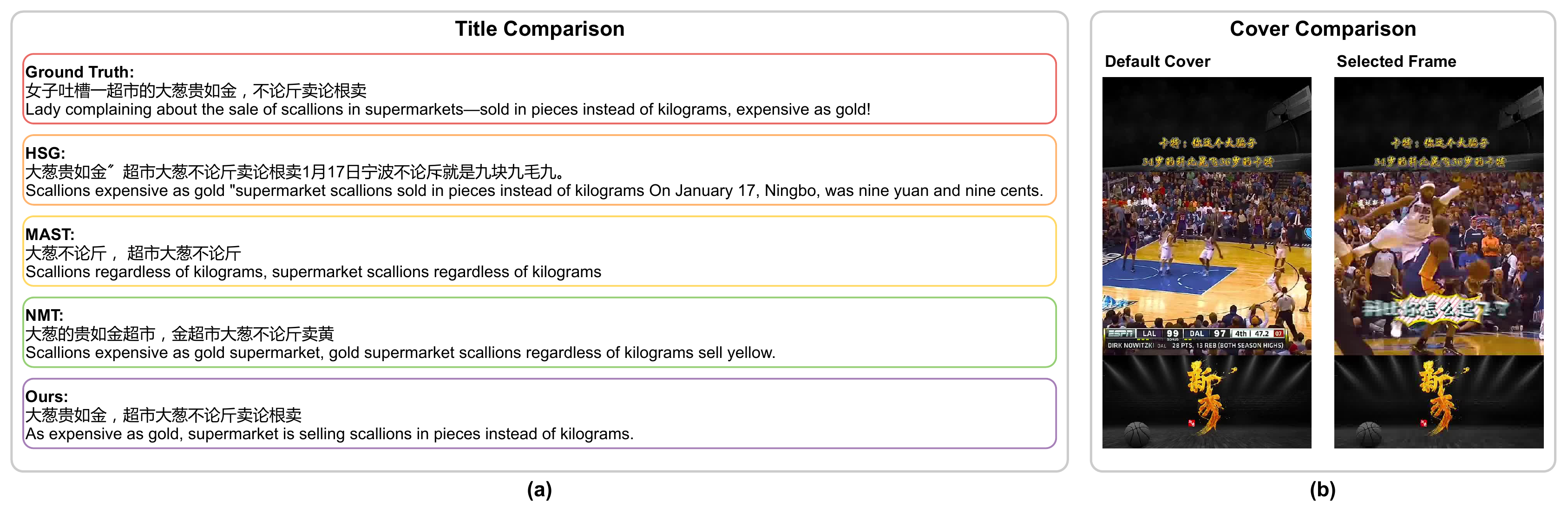}
		\caption{Examples of (a) the generated titles by the baselines and TCR (b) the selected frame and its default cover that comes with the video. The video title is ``Kobe Bryant pump fakes Vince Carter''.}
	\label{fig:case}
	\vspace{-3mm}
\end{figure*}

Fig. \ref{fig:case} shows examples including the generated titles by baseline models and our proposed model, as well as the cover selection comparison.
The results from Fig. \ref{fig:case}(a) demonstrate that our proposed model could generate a more accurate title than the baselines. 
Furthermore, as shown in Fig. \ref{fig:case}(b), the selected frame is better than the default cover that comes with the video since the selected frame captures the jumping moment of the defence player while the default cover does not. To better understand the refinement process and what our model has learned, we also visualize the attention scores in Supplementary G.

\section{Conclusion}
In this paper, we first collect a real-world dataset named SVTG to specifically support the joint TG-CS task for short videos. To our best knowledge, SVTG is the first short video dataset that contains appealing titles instead of unvarnished factual captions or summaries like other benchmark datasets. 
Then we propose a TCR method that consists of a title-cover generator and a model refinement module. 
The title-cover generator generates the title and extracts the cover image by fully exploiting the dependency between modalities. 
The refinement module further selects high-quality training samples and relevant text/frames within each sample to refine the generator progressively.
Experiments on the SVTG dataset and the public How2 dataset show the effectiveness of the proposed TCR method on title generation when given noisy data.
Furthermore, the cover selection evaluation results suggest that our selected covers are preferred compared to the default covers. This further demonstrates the effectiveness of the TCR method on cover selection.

\bibliographystyle{splncs04}
\bibliography{mybibliography}

\end{document}